\documentclass[11pt]{article}
\usepackage{coling2020}
\usepackage{times}
\usepackage{url}
\usepackage{latexsym}

\colingfinalcopy % Uncomment this line for the final submission

\usepackage{calc}
\usepackage{ifthen}
\usepackage{tikz}
\usepackage{url} 
\usepackage{graphicx} 
\usepackage{tabularx}
\usepackage{footnote}\makesavenoteenv{tabular}
\usepackage{float}
\usepackage{rotating,booktabs,multirow}
\usepackage{adjustbox}
\usepackage{scalerel,amssymb, amsmath}
\usepackage[nointegrals]{wasysym}
\usepackage{textcomp}
\def\mcirc{\mathbin{\scalerel*{\circ}{j}}}
\def\bcirc{\mathbin{\scalerel*{\CIRCLE}{j}}}

\usepackage{CJKutf8}

\usepackage{array}
\newcolumntype{L}{>{\arraybackslash}m{\linewidth}}

\usepackage{xcolor}

\title{A BERT-based Dual Embedding Model for\\Chinese Idiom Prediction}

\author{Minghuan Tan \\
  School of Information Systems \\
  Singapore Management University \\
  \texttt{mhtan.2017@phdcs.smu.edu.sg} \\\And
  Jing Jiang \\
  School of Information Systems \\
  Singapore Management University \\
  \texttt{jingjiang@smu.edu.sg} \\}

\date{}

\begin{document}
\maketitle

% \jjcomment{Add author names.}

\begin{abstract}
Chinese idioms are special fixed phrases usually derived from ancient stories, whose meanings are oftentimes highly idiomatic and non-compositional.
% Chinese idioms should be be used in consistent with the context both semantically and syntactically.
The Chinese idiom prediction task is to select the correct idiom from a set of candidate idioms given a context with a blank.
We propose a BERT-based dual embedding model to encode the contextual words as well as to learn dual embeddings of the idioms.
Specifically, we first match the embedding of each candidate idiom with the hidden representation corresponding to the blank in the context.
We then match the embedding of each candidate idiom with the hidden representations of all the tokens in the context thorough context pooling. 
We further propose to use two separate idiom embeddings for the two kinds of matching.
Experiments on a recently released Chinese idiom cloze test dataset show that our proposed method performs better than the existing state of the art.
Ablation experiments also show that both context pooling and dual embedding contribute to the improvement of performance.
\end{abstract}
\begin{CJK*}{UTF8}{gbsn}

\section{Introduction}
\label{sec:intro}

%
% The following footnote without marker is needed for the camera-ready
% version of the paper.
% Comment out the instructions (first text) and uncomment the 8 lines
% under "final paper" for your variant of English.
% 
\blfootnote{
    %
    % for review submission
    %
    % \hspace{-0.65cm}  % space normally used by the marker
    % Place licence statement here for the camera-ready version. See
    % Section~\ref{licence} of the instructions for preparing a
    % manuscript.
    %
    % % final paper: en-uk version 
    %
    % \hspace{-0.65cm}  % space normally used by the marker
    % This work is licensed under a Creative Commons 
    % Attribution 4.0 International Licence.
    % Licence details:
    % \url{http://creativecommons.org/licenses/by/4.0/}.
    % 
    % final paper: en-us version 
    
    \hspace{-0.65cm}  % space normally used by the marker
    This work is licensed under a Creative Commons 
    Attribution 4.0 International License.
    License details:
    \url{http://creativecommons.org/licenses/by/4.0/}.
}

In this paper, we study Chinese idiom prediction, a language understanding problem that has not been extensively explored before in computational linguistics.
Chinese idioms, mainly Chengyu~(成语)~(set phrases)~\cite{wang-yu-2010-construction,wang2019chinese}, have fixed forms in structure; the component characters~(mostly four) cannot be changed.
% In literature, Chinese Idioms are the focus of Chinese lexicology research.  
Chinese idioms are characterized by rich contents, concise forms and frequent use~\cite{wang2019chinese} with properties of structural regularity, semantic fusion, and functional integrity~\cite{shao2018,wang2019chinese}. 
Chinese idioms are commonly used in both written and spoken Chinese, and understanding Chinese idioms is important for learning Chinese as a second language.

The meaning of each Chinese idiom may not be literally understood through the composition of its characters, especially for those which are derived from historical stories or formulated using ancient Chinese grammars.
For example, ``一定不易" is literally interpreted as \emph{``it must be not easy''} in modern Chinese.
However, the idiom is constructed from grammars and word senses of ancient Chinese.
Its idiomatic meaning is \emph{``once decided, never change''}, which is not even close to the literal meaning.
As a result, the usage of Chinese idioms poses a challenge on language understanding not only for humans but also for artificial intelligence. 
Due to their pervasive usage, Chinese idiom prediction is an important task in Chinese language understanding.

% \mhresponse{As requested by the reviewers, we need to update the translation and make the examples consistent in this paper. Therefore, I will replace the table as following:}

\begin{table}[ht]
    \centering
    \begin{adjustbox}{max width=\linewidth}
    % \small
    \begin{tabular}{L}
        \toprule
        \textbf{Passage:}
        戴尔克·施特略夫把自己的工作全部撂下，整天服侍病人，又体贴，又关切。他的手脚非常利索，把病人弄得舒舒服服。大夫开了药，他总是连哄带骗地劝病人按时服用，我从来没想到他的手段这么巧妙。无论做什么事他都不嫌麻烦。尽避他的收入一向只够维持夫妻两人的生活，从来就不宽裕，现在他却\underline{\hspace{1cm}}，购买时令已过、价钱昂贵的美味，想方设法叫思特里克兰德多吃一点东西（他的胃口时好时坏，叫人无法捉摸）。\\
        Dirk Stroeve, giving up his work entirely, nursed Strickland with tenderness and sympathy. He was dexterous to make him comfortable, and he exercised a cunning of which I should never have thought him capable to induce him to take the medicines prescribed by the doctor. Nothing was too much trouble for him. Though his means were adequate to the needs of himself and his wife, he certainly had no money to waste; but now he was \underline{\hspace{1cm}} in the purchase of delicacies, out of season and dear, which might tempt Strickland’s capricious appetite. \\
        \midrule
            \textbf{Candidates:} \\
            $\mcirc$ 月明星稀~The moon is bright and stars are few; with a clear moon and few stars \\
            $\mcirc$ 苦尽甘来~bitterness ends and happiness begins  \\
            $\mcirc$ 坐吃山空~even a great fortune can be depleted by idleness \\
            $\bcirc$ 大手大脚~extravagant or wasteful \\
            $\mcirc$ 斤斤计较~haggle over every ounce  \\
            $\mcirc$ 不见天日~a world of darkness; total absence of justice\\
            $\mcirc$ 好吃懒做~be fond of eating and averse to work; be gluttonous and lazy \\
        \bottomrule
    \end{tabular}
    \end{adjustbox}
    \caption{An example showing a passage with a blank and seven candidate idioms. The idiom with the solid circle is the ground truth idiom. The passage is from a Chinese translation of \textit{The Moon and Sixpence}. Translations of idioms are extracted from online dictionary \url{http://dict.cn}.}
    \label{tab:example}
\end{table}

% ["月明星稀", "苦尽甘来", "坐吃山空", "大手大脚", "斤斤计较", "不见天日", "好吃懒做"]
% 戴尔克·施特略夫把自己的工作全部撂下，整天服侍病人，又体贴，又关切。他的手脚非常利索，把病人弄得舒舒服服。大夫开了药，他总是连哄带骗地劝病人按时服用，我从来没想到他的手段这么巧妙。无论做什么事他都不嫌麻烦。尽避他的收入一向只够维持夫妻两人的生活，从来就不宽裕，现在他却大手大脚，购买时令已过、价钱昂贵的美味，想方设法叫思特里克兰德多吃一点东西（他的胃口时好时坏，叫人无法捉摸）。我什么时候也忘不了他劝说思特里克兰德增加营养的那种耐心和手腕。不论思特里克兰德对他多么没礼貌，他也从来不动火。如果对方只是郁闷懊丧，他就假装看不到；如果对方顶撞他，他只是一笑置之。当思特里克兰德身体好了一些，情绪高起来，嘲笑他几句开开心，他就做出一些滑稽的举动来，故意给对方更多讥笑的机会。他会高兴地递给我几个眼色，叫我知道病人已经大有起色了。施特略夫实在是个大好人。

% Dirk Stroeve, giving up his work entirely, nursed Strickland with tenderness and sympathy. He was dexterous to make him comfortable, and he exercised a cunning of which I should never have thought him capable to induce him to take the medicines prescribed by the doctor. Nothing was too much trouble for him. Though his means were adequate to the needs of himself and his wife, he certainly had no money to waste; but now he was wantonly extravagant in the purchase of delicacies, out of season and dear, which might tempt Strickland’s capricious appetite. I shall never forget the tactful patience with which he persuaded him to take nourishment. He was never put out by Strickland’s rudeness; if it was merely sullen, he appeared not to notice it; if it was aggressive, he only chuckled. When Strickland, recovering somewhat, was in a good humour and amused himself by laughing at him, he deliberately did absurd things to excite his ridicule. Then he would give me little happy glances, so that I might notice in how much better form the patient was. Stroeve was sublime.

There have been several studies focusing on representing Chinese idioms using neural network models~\cite{jiang-etal-2018-chengyu,liu-etal-2019-neural}, but they were limited by the amount of data available for training.
Recently, Zheng et al.~\shortcite{zheng-etal-2019-chid} released a large-scale Chinese IDiom Dataset~(ChID) to facilitate machine comprehension of Chinese idioms.
The ChID dataset contains more than 500K passages and 600K blanks, making it possible for researchers to train deep neural network models.
The dataset is in cloze test style that target Chinese idioms in passages are replaced by blanks.
For each blank, a set of candidate Chinese idioms is provided and the task is to pick the correct one based on the context.
Table~\ref{tab:example} shows an example from the testing set of ChID.
We can see that among the seven candidates, most can fit into the local context ``现在他却\underline{\hspace{1cm}}'' (``but now he was \underline{\hspace{1cm}}'') well grammatically, but to select the best answer we need to understand the entire passage.

In this paper, we propose a BERT-based dual embedding model for the Chinese idiom prediction task.
We first present two baseline models that use BERT to process and match passages and candidate answers in order to rank the candidates.
Observing that these baselines do not explicitly model the global, long-range contextual information in the given passage for Chinese idiom prediction, we propose a context-aware pooling operation to force the model to explicitly consider all contextual words when matching a candidate idiom with the passage.
Furthermore, we propose to split the embedding vector of each Chinese idiom into two separate vectors, one modeling its local properties and the other modeling its global properties.
We expect the embedding for local properties to capture the syntactic properties of an idiom, while the embedding for global properties to capture its topical meaning.
In addition, using idiom embeddings makes it possible for us to consider the entire Chinese idiom vocabulary as the candidate set, which is computationally intractable compared to pretrained BERT models with multiple-sequence classification. we apply this enlarged candidates heuristic to all the models with idiom embeddings to further strengthen the performance.

To evaluate the effectiveness of the BERT-based dual embedding model, we conduct experiments on the ChID dataset.
Our experiments show that our method can outperform several existing methods tested by~\newcite{zheng-etal-2019-chid} as well as our baseline methods. 
We also find that both context-aware pooling and dual embedding contribute to the performance improvement.
To prove the effectiveness of our model, we also evaluate it against a public leaderboard of ChID Competition.
The results show that our model is competitive compared to the top-ranked systems.
We can also achieve better performance with a large margin compared with several methods using pretrained language models.
% with multiple-sequence classification.
We also conduct further analysis using a gradient-based attribution method to check if our model can indeed capture global information to make correct predictions.
Some case studies show that indeed our method makes use of more global contextual information to make predictions.
\section{Related Work}

\subsection{Cloze-style Reading Comprehension} 

Cloze-style reading comprehension is an important form in assessing machine reading abilities.
Researchers created many large-scale cloze-style reading comprehension datasets like CNN/Daily Mail~\cite{NIPS2015_5945}, Children's Book Test~(CBT)~\cite{hill2015goldilocks} and RACE~\cite{lai-etal-2017-race}.
These datasets have inspired the design of various neural-based models~\cite{NIPS2015_5945,chen-etal-2016-thorough} and some become benchmarks for machine reading comprehension.
The dataset ChID used in this paper is also a large scale cloze-style dataset but focuses on Chinese idiom prediction.

\subsection{Pre-trained Language Models} 

Language model pre-training has been proven to be effective over a list of natural language tasks at both sentence level~\cite{bowman-etal-2015-large} and token level~\cite{tjong-kim-sang-de-meulder-2003-introduction,rajpurkar-etal-2016-squad}.
Existing strategies of using pre-trained language models include feature-based methods like ELMO~\cite{peters-etal-2018-deep} and fine-tuning methods such as OpenAI GPT~\cite{radford2018improving} and BERT~\cite{devlin-etal-2019-bert}.
BERT-based fine-tuning strategy and its extensions~\cite{cui2019pre,NIPS2019_8812,liu2019roberta} are pushing performance of neural models to near-human or super-human level.
In this paper, we use pre-trained Chinese BERT with Whole Word Masking~\cite{cui2019pre} as text sequence processor.

\subsection{Modelling Figurative Language} 
Figurative (or non-literal) language is different from literal language where words or characters in literal language act in accordance with  conventionally accepted meanings or denotation. 
In figurative language, meaning can be detached from the words or characters while a more complicated meaning or heightened effect is reattached.
As a special type of figurative language, idioms have been actively researched in tasks like Idiom Identification~\cite{muzny-zettlemoyer-2013-automatic}, Idiom Recommendation~\cite{liu-etal-2019-neural} and Idiom Representation~\cite{gutierrez-etal-2016-literal,liu-etal-2017-idiom,jiang-etal-2018-chengyu,zheng-etal-2019-chid}.
In this paper, we will focus on the representations of Chinese idioms using a BERT-based approach.
\section{Method}
\label{sec:method}

% The Chinese idiom prediction task we study can be regarded as a standard cloze-style reading comprehension problem, which has been extensively studied in recent years in the natural language processing and machine learning communities~\cite{cnn,cloze_teacher}.
% In~\cite{chid}, the authors showed the performance of a number of obvious baselines for the task on the \textbf{ChID} dataset.
% The baselines they considered are the following:
% \textbf{Language Model:} This method is based on standard bidirectional LSTM~(BiLSTM)~\cite{lstm,bilstm}. It uses BiLSTM to encode the given passage and compares it with the embedding vector of each candidate idiom in order to select the best idiom.
% \textbf{Attentive Reader:} This method also uses BiLSTM but augments it with attention mechanism to obtain an attention-weighted encoding of the passage.
% It is based on the Attentive Reader model by~\cite{ar}.
% \textbf{Standard Attentive Reader:} This is an altered version of Attentive Reader, where attention weights are computed using a bilinear matrix.
% It is based on~\cite{sar}.

\subsection{Task Definition and Dataset}

We formally define the Chinese idiom prediction task as follows.
Given a passage $P$, represented as a sequence of tokens $(p_1, p_2, \ldots, p_n)$, where each token is either a Chinese character or the special ``blank'' token \texttt{[MASK]}, and a set of $K$ candidate Chinese idioms denoted as $\mathcal{A} = \{a_1, a_2, \ldots, a_K\}$, 
our goal is to select an idiom $a^* \in \mathcal{A}$ that best fits the blank in $P$.
See the example in Table~\ref{tab:example}.

We assume that a set of training examples in the form of triplets, each containing a passage, a candidate set and the ground truth answer, is given.
We denote the training data as $\{(P_i, \mathcal{A}_i, a^*_i)\}_{i=1}^N$.
We use $\mathcal{V}$ to denote the vocabulary of all Chinese idioms observed in the training data, i.e., $\mathcal{V} = \cup_{i=1}^N \mathcal{A}_i$.

To facilitate the study of Chinese idiom comprehension using deep learning models, Zheng et al.~\shortcite{zheng-etal-2019-chid} released the ChID dataset.
The dataset was created in the ``cloze'' style.
The authors collected diverse passages from novels and essays on the Internet and news articles from THUCTC~\cite{THUCTC}.
The authors then replaced target Chinese idioms found in these passages with the blank token.
To construct the candidate answer set for each blank, the authors considered synonyms, near-synonyms and other idioms either irrelevant or opposite in meaning to the ground truth idiom~\cite{zheng-etal-2019-chid}.

\subsection{BERT Baselines}
\label{subsec:baselines}

Previous methods applied to the ChID dataset are not based on BERT~\cite{devlin-etal-2019-bert} or Transformer~\cite{NIPS2017_7181} architecture.
Because of the success of BERT for many NLP tasks, here we first present two BERT baselines. 
The first one treats a Chinese idiom as a sequence of characters.
It combines the passage with each candidate idiom into a single sequence and processes multiple sequences, one for each candidate, using BERT.
The second one treats a Chinese idiom as a single token that has its own embedding vector.
The method uses BERT to process the passage and then matches the encoded passage with each candidate idiom's embedding.
These baselines can be regarded as standard ways to solve the Chinese idiom prediction problem using BERT.

For the second baseline that uses idiom embeddings, we also present a heuristic that uses an enlarged candidate set to improve learning.
This heuristic is only applicable to the second baseline because it would be computationally too expensive for the first baseline. 

\paragraph{BERT Baseline with Idioms as Character Sequences:}

A straightforward way to apply BERT for Chinese idiom prediction is as follows.
Given a passage $P = (p_1, p_2, \ldots, \texttt{[MASK]}, \ldots, p_n)$ and a candidate answer $a_k \in \mathcal{A}$, we first concatenate them into a single sequence
$(\texttt{[CLS]}, p_1, p_2, \ldots, p_n, \texttt{[SEP]},  a_{k, 1}, a_{k, 2},  a_{k, 3}, \allowbreak a_{k, 4},\texttt{[SEP]})$, where $a_{k, 1}$ to $a_{k, 4}$ are the four Chinese characters that idiom $a_k$ is composed of.
We can then directly use BERT to process this sequence and obtain the hidden representation for $\texttt{[CLS]}$ on the last ($L$-th) layer, denoted by $\mathbf{h}^L_{k, 0} \in \mathbb{R}^d$.
To select the best answer idiom, we first use a linear layer to process $\mathbf{h}^L_{k, 0}$ for $k = 1, 2, \ldots, K$ and then use standard softmax to obtain the probabilities of each candidate.
To train the model, we use standard negative log likelihood as the loss function.

\paragraph{BERT Baseline with Idiom Embeddings:}
Many Chinese idioms are non-compositional and therefore their meanings should not be directly derived from the embeddings of its four individual characters, as the baseline above does. 
E.g., ``狐假虎威'' literally means a fox assuming the majesty of a tiger, but it is usually used to describe someone flaunting his powerful connections.
Therefore, we hypothesize that learning a single embedding vector for the entire idiom can help the understanding of idioms.

In this second BERT baseline, instead of concatenating the passage and a candidate answer into a single sequence for BERT to process, we keep them separated.
We only use BERT to process the passage sequence $(\texttt{[CLS]}, p_1, p_2, \ldots, \texttt{[MASK]}, \ldots, p_n, \texttt{[SEP]})$.
Afterwards, we use the hidden representation of $\texttt{[MASK]}$ at the last ($L$-th) layer, denoted as $\mathbf{h}^L_b$, to match each candidate answer.
In this way, no matter how many candidate answers there are, BERT is used to process the passage only once.
On the other hand, each Chinese idiom has a hidden embedding vector, which is to be learned.

We use $\mathbf{a}_k$ to denote the embedding vector for candidate $a_k \in \mathcal{A}$.
The hidden representation $\mathbf{h}^L_b$ is fused with each candidate idiom via element-wise multiplication.
Then the probability of selecting $a_k$ among all the candidates $\mathcal{A}$ is defined as follows:
\begin{equation}
    p_k = \frac{\exp(\mathbf{w}\cdot(\mathbf{a}_k \odot \mathbf{h}^L_b)+b)}{\sum_{k'=1}^K\exp(\mathbf{w}\cdot(\mathbf{a}_{k'} \odot \mathbf{h}^L_b) + b)},
        \label{eqn:p_k_baseline}
\end{equation}
where $\mathbf{w} \in \mathbb{R}^d$ and $b \in \mathbb{R}$ are model parameters, and $\odot$ is element-wise multiplication.
To train the model, we again use negative log likelihood as the loss function.

\paragraph{Heuristic with Enlarged Candidate Set:}

The ChID dataset uses only a small set of negative answers in each candidate set and these negatives are fixed for each example during training.
It is reasonable to expect that most of the remaining Chinese idioms not in the candidate set are also negative answers and including them in the training data may help.
We therefore use a heuristic that considers an enlarged candidate set to further boost the performance.

To apply this heuristic, we define a candidate set $\mathcal{A}'$ to be the same as $\mathcal{V}$ (i.e., the vocabulary containing all Chinese idioms observed in the training data), and then define a second term in the loss function that is the negative log likelihood of selecting the correct answer from this enlarged candidate set.

Note that because $\mathcal{A}'$ is large, this heuristic is not feasible to be applied to the character sequence-based BERT baseline, because it would require inserting each candidate into the passage for BERT to process, which would be computationally too expensive. 
Therefore, this enlarged candidate set heuristic is only applied to the idiom embedding-based BERT baseline.
Specifically, we can define the probability of selecting answer $a \in \mathcal{A}'$ as follows:
\begin{equation}
    q_{a} = \frac{\exp(\mathbf{a}\cdot\mathbf{h}^L_b)}{\sum_{c \in \mathcal{A}'} \exp(\mathbf{c}\cdot\mathbf{h}^L_b)}.
\end{equation}
Let $q^*_i$ denote the probability of selecting the ground truth idiom among all candidates in $\mathcal{A}'$ for the $i$-th training example, and $p^*_i$ denote the probability of selecting the correct answer among the original candidate set $\mathcal{A}$ for the $i$-th training example.
Our training loss function is then defined as follows:
\begin{equation}
    L = -\sum_{i=1}^N (\log(p^*_i) + \log(q^*_i)).
\end{equation}

\subsection{Our Dual Embedding Model}
\label{subsec:our_model}

The BERT baselines presented above are reasonable baselines, but they have a potential problem.
% \jjcomment{We have to stress this problem with the baselines in the introduction section. We also need to point out that previous methods for this task didn't address this problem, either.}
% (1) As we can see, our goal is to evaluate whether a candidate answer $a_k$ can be used to replace the blank token $\texttt{[MASK]}$ such that the completed passage is coherent, so it is important to test whether $a_k$ can well fit into the local context surrounding $\texttt{[MASK]}$.
% But the baseline models above do not allow us to place more emphasis on the interactions between $a_k$ and the tokens surrounding $\texttt{[MASK]}$.
% \jjcomment{I find this argument above not convincing. How does our dual embedding model eanble the interactions between $a_k$ and the tokens surrounding $\texttt{[MASK]}$? I feel we need to remove this argument.}
% (2) 
We observe that in order for an idiom to fit into a passage well, it has to not only grammatically (i.e., syntactically) fit into the local context surrounding the $\texttt{[MASK]}$ token but also show semantic relevance to the whole passage.
In the example shown in Table~\ref{tab:example}, a correct answer has to first be an adjective rather than, say, a noun or a verb.
In addition, given the global context of the entire passage, it is understood that the correct answer should convey the meaning of ``extravagant.''

Based on the observation above, we introduce the following two changes to the second BERT baseline, i.e., the idiom embedding-based BERT baseline, introduced in Section~\ref{subsec:baselines}.

\subsubsection{Context-aware Pooling}
\label{subsubsec:context-pooling}

As we have pointed out earlier, oftentimes Chinese idioms have non-compositional meanings, and to evaluate whether a Chinese idiom is suitable in a passage, we need to understand the semantic meaning of the entire passage.
Therefore, it is important for us to not only try to match an idiom with the local context it is to be placed in (which can roughly be modeled by $\mathbf{h}^L_b$) but also to match it with the entire passage.
Let us use $\mathbf{a}_k$ to denote the embedding for idiom $a_k$.
Recall that $\mathbf{H}^L = (\mathbf{h}^L_0, \mathbf{h}^L_1, \ldots, \mathbf{h}^L_n)$ represents the hidden states of the last layer of BERT after it processes the passage sequence.
Our method with context-aware pooling can be represented as follows:
\begin{equation}
    p_k = \frac{\exp(\mathbf{a}_k\cdot\mathbf{h}^L_b + \max_{i = 0}^n (\mathbf{a}_k \cdot \mathbf{h}^L_i))}{\sum_{k'=1}^K\exp(\mathbf{a}_{k'}\cdot\mathbf{h}^L_b + \max_{i = 0}^n (\mathbf{a}_{k'} \cdot \mathbf{h}^L_i))}.
\end{equation}

\subsubsection{Dual Embeddings}
\label{subsubsec:dual}

% \jjcomment{I removed the original text here because I now realized that you're not matching the idiom embedding with $\mathbf{h}^L_0$.}
% (which can be modeled by $\mathbf{h}^L_0$, the hidden state corresponding to the special token $\texttt{[CLS]}$).
% Note that it has been common practice when applying BERT to use the hidden state corresponding to $\texttt{[CLS]}$ for sequence-level predictions~\cite{bert,xlnet,roberta}. 

Because we need to match an idiom with both $\mathbf{h}^L_b$ and the entire passage, the second idea we propose is to split the embedding of an idiom into two ``sub-embedding'' vectors, which we refer to as ``dual embeddings.''
Let us use $\mathbf{a}^u_k$ and $\mathbf{a}^v_k$ to denote the two embeddings for idiom $a_k$.

We then calculate the probability of selecting candidate $a_k$ as follows:
\begin{equation}
    p_k = \frac{\exp(\mathbf{a}_k^u\cdot\mathbf{h}^L_b + \max_{i = 0}^n (\mathbf{a}_k^v \cdot \mathbf{h}^L_i))}{\sum_{k'=1}^K\exp(\mathbf{a}_{k'}^u\cdot\mathbf{h}^L_b + \max_{i = 0}^n (\mathbf{a}_{k'}^v \cdot \mathbf{h}^L_i))}.
\end{equation}

We also adopt the heuristic of enlarged candidate set from Section~\ref{subsec:baselines}.
With the candidate set $\mathcal{A}'$ to be the same as $\mathcal{V}$, we still use dual embeddings to represent each idiom, but when we match the dual embeddings with the passage, we use both $\mathbf{a}^u$ and $\mathbf{a}^v$ to match $\mathbf{h}^L_b$ only.
This is because it would be too expensive to match $\mathbf{a}^v$ of each candidate with the entire sequence of hidden states $\mathbf{H}^L$ as we now have many candidates.
So we define the probability of selecting answer $a \in \mathcal{A}'$, i.e., selecting the ground truth answer from the entire vocabulary of Chinese idioms, as follows:
\begin{equation}
    q_{a} = \frac{\exp(\mathbf{a}^u\cdot\mathbf{h}^L_b+\mathbf{a}^v\cdot\mathbf{h}^L_b)}{\sum_{c \in \mathcal{A}'} \exp(\mathbf{c}^u\cdot\mathbf{h}^L_b+\mathbf{c}^v \cdot \mathbf{h}^L_b)}.
\end{equation}

Similarly, to train the model, we use negative log likelihood as shown before.

\section{Experiments}

% In this section, we first evaluate our proposed dual embedding method using the ChID Official dataset.
% We will compare the results of our proposed method with the models of earlier literature as well as the two BERT baselines presented earlier.
% Then we report our performance on the leaderboard of ChID Competition against several high ranked systems to further illustrate the competency of our method.

In this section, we evaluate our proposed dual embedding method using the ChID dataset.
% \textbf{ChID-Official} dataset and compare the results with those of existing methods as well as the two BERT baselines.
% Then we report our performance on the leaderboard of \textbf{ChID-Competition} alongside the performance of highly ranked systems to further illustrate the effectiveness of our method.
% Finally, 
We also use an attribution method to visualize how each proposed method works on some selected cases.

\subsection{Evaluation on ChID-Official}
% Experiment Settings

\begin{table*}[th]
\centering
\begin{adjustbox}{max width=\linewidth}
% \small
\begin{tabular}{lrrrrrr}
    \toprule
        % Model & Dev & Test & Ran & Sim & Out & Test-MRR\\
    \multirow{2}[4]{*}{} & \multicolumn{4}{c}{In-domain} & Out-of-domain & Total\\ 
    \cmidrule(rl){2-5} \cmidrule(rl){6-6} \cmidrule(rl){7-7} & Train  & Dev   & Test  & Total  & Out  & Total  \\         
    \midrule
        Passages   &  520,711	&20,000	&20,000	&560,711	&20,096	&580,807  \\
        Distinct idioms &	3,848	&3,458&	3,502	&3,848	&3,626	&3,848 \\
        Total blanks	&648,920	&24,822	&24,948	&698,690	&30,023	&728,713\\
    \bottomrule
\end{tabular}
\end{adjustbox}
\caption{Some statistics of the \textbf{ChID} dataset.}
\label{tab:data}
\end{table*}    

\paragraph{Data Split:} 
In the first set of experiments, We use the official release of ChID\footnote{\url{https://github.com/zhengcj1/ChID-Dataset}}, denoted as \textbf{ChID-Official}.
The data has a training set, a development set and a few different test sets.
Besides the standard test set \textbf{Test}, the authors also constructed the following test sets:
\textbf{Ran}: In this test set, the candidate idioms are randomly sampled from the vocabulary $\mathcal{V}$. No synonyms or near-synonyms were intentionally added as candidates.
\textbf{Sim}: In this test set, the candidates are sampled from the top-10 similar idioms and are more challenging than the Ran test dataset.
The only difference of \textbf{Test}, \textbf{Ran} and \textbf{Sim} is the candidate sets.
\textbf{Out}: This is an out-of-domain test dataset. The passages come from essays (whereas the training and development data comes from news and novels).
Statistics of the data can be found in Table~\ref{tab:data}.
% \jjcomment{Does Table 2 show the statistics of the ChID-Official dataset? We mentioned Ran and Sim here but they're not found in the table, which may cause confusion. The best is to make the table about dataset statistics consistent with our description here. Or, we should give some explanation. The bottom line is to avoid any confusion.}

\paragraph{Methods Compared:}

We compare the following different methods.
Performance of the first three baselines are directly taken from~\cite{zheng-etal-2019-chid}.
It is worth noting that the three baselines use BiLSTM as their backbones while our methods use BERT (Transformer) as our backbones.
Although BiLSTM with attention can also capture the global contextual information in the passages, our experiments below will show that empirically our BERT-based methods are more effective.

\textbf{Language Model~(LM):} This method is based on standard bidirectional LSTM~(BiLSTM)~\cite{lstm,zhou-etal-2016-attention-based}. It uses BiLSTM to encode the given passage and obtain the hidden state of the blank.
Then it compares the blank state with the embedding vector of each candidate idiom to choose the best idiom.

\textbf{Attentive Reader~(AR):} This method also uses BiLSTM but augments it with attention mechanism. 
It is based on the Attentive Reader model by~\cite{NIPS2015_5945}.

\textbf{Standard Attentive Reader~(SAR):} This is an altered version of Attentive Reader, where attention weights are computed using a bilinear matrix~\cite{chen-etal-2016-thorough}.
% as described in Section~\ref{subsec:baselines}.

\textbf{BL-CharSeq:} This is the first BERT baseline treating idioms as character sequences.
% of multiple-sequence classification as described in Section~\ref{subsec:bert-base}.
% We refer to this baseline as \textbf{BL-CharSeq}.

\textbf{BL-IdmEmb~(w/o EC):} This is the second BERT baseline using idiom embeddings.
In this version, we do not use enlarged candidate set.

\textbf{BL-IdmEmb:} This baseline is the same as BL-IdmEmb~(w/o EC) but incorporates the heuristic of enlarged candidate set.

\textbf{Ours-CP:} This is our method with contextual pooling~(CP) as presented in Section~\ref{subsubsec:context-pooling}. 
This method also incorporates the enlarged candidate set heuristic.

\textbf{Ours-Full~(CP+DE):} This is our method with both context pooling~(CP) and dual embedding~(DE), as presented in Section~\ref{subsubsec:dual}.
This method also uses the enlarged candidate set heuristic.

\paragraph{Evaluation Metrics:}
A standard metric for the task of Chinese idiom prediction is accuracy, which is the percentage of test examples where our predicted idiom is the same as the ground truth idiom.
Here besides accuracy, we also consider another setting where we do not have a pre-defined set of candidate idioms, or in other words, we consider \emph{all} Chinese idioms in our vocabulary as candidates.
% \textcolor{blue}{Based on the similarity between idiom embeddings and the hidden state of the blank, we report the performance of ranking all the Chinese idioms in the vocabulary.
For this setting, we use Mean Reciprocal Rank~(MRR)~\cite{Voorhees99thetrec-8,radev-etal-2002-evaluating}, a well-established metric for ranking problems, as the evaluation metric.

\paragraph{Other Settings:}
We use pre-trained BERT for Chinese with Whole Word Masking~(WWM)~\cite{cui2019pre}\footnote{\url{https://github.com/ymcui/Chinese-BERT-wwm}}.
% \jjcomment{This part is confusing. For our methods, which are based on pre-trained BERT, do we use BERT with WWM or do we use RoBERTa for Chinese? Also, anything related to Table 4 should be mentioned later in the next subsection, so if RoBERTa is only used because of the comparison in Table 4, let's mention RoBERTa only in the next subsection.}
% Due to computationally reasons, 
To reduce computational cost,
we choose 128 as the maximum length for the input sequence, and we truncate passages longer than this limit by keeping only the 128 characters surrounding $\texttt{[MASK]}$, with $\texttt{[MASK]}$ in the middle.

We use 4 Nvidia 1080Ti GPU cards and a batch size of 10 per card with a total 5 training epochs.
The initial learning rate is set to $5e^{-5}$ with 1000 warm-up steps.
We use the optimizer \textit{AdamW} in accordance with a learning rate scheduler \textit{WarmupLinearSchedule}.
Our code has been made available online\footnote{\url{https://github.com/VisualJoyce/ChengyuBERT}}.

\begin{table*}[t]
\centering
\begin{adjustbox}{max width=\linewidth}
% \small
\begin{tabular}{lcccccccccc}
    \toprule
        % Model & Dev & Test & Ran & Sim & Out & Test-MRR\\
    \multirow{2}[4]{*}{} & \multicolumn{2}{c}{Dev} & \multicolumn{2}{c}{Test} & \multicolumn{2}{c}{Ran} & \multicolumn{2}{c}{Sim} & \multicolumn{2}{c}{Out}\\ 
    \cmidrule(rl){2-3} \cmidrule(rl){4-5} \cmidrule(rl){6-7}\cmidrule(rl){8-9}\cmidrule(rl){10-11} & ACC  & MRR   & ACC  & MRR  & ACC  & MRR  & ACC  & MRR  & ACC  & MRR  \\         
    \midrule
        \makebox[0pt][l]{Human}\phantom{Ours-Single}~\cite{zheng-etal-2019-chid}          & -& - & 87.1& - &97.6 & -&82.2& -& 86.2& -  \\
    \midrule
        \makebox[0pt][l]{LM}\phantom{Ours-Single}~\cite{zheng-etal-2019-chid}        & 71.8& - & 71.5& - &80.7& - &65.6& -& 61.5& - \\
        \makebox[0pt][l]{AR}\phantom{Ours-Single}~\cite{zheng-etal-2019-chid}        & 72.7& - & 72.4& - & 82.0& - &66.2& - &62.9& -\\
        \makebox[0pt][l]{SAR}\phantom{Ours-Single}~\cite{zheng-etal-2019-chid}         & 71.7& - & 71.5& - & 80.0& - &64.9& - & 61.7& -\\
    \midrule
        BL-CharSeq	& 79.33	& -&79.42 & - &88.84	&-&72.93&-	&73.11&-	\\
        BL-IdmEmb~(w/o EC)	& 73.59 & 0.017&73.31 & 0.017 &81.05	&0.017&68.13&0.017	&63.82&0.012	\\
    %     Ours-Single	& 79.08	& 0.296 & 79.20 & 0.295 & 87.37	& 0.295 &73.61& 0.295&70.64& 0.200	\\
    %     Ours-Dual	& 80.69	& 0.290 & 80.56 & 0.287 & 88.62	& 0.287 & 74.74 & 0.287		& 73.36 & 0.201	\\        
    %     % Ours-Single & 80.11	& 0.277 & 79.98	 & 0.275 &88.52	 & 0.275 &74.39	 & 0.275 &72.78	& 0.188 \\        
    %     % Ours-Dual & 81.15	& 0.286 & 81.04	 & 0.282 &89.27	 & 0.282 &\textbf{75.69}	 & 0.282 &74.06	& 0.194 \\     
    % \midrule
        \makebox[0pt][l]{BL-IdmEmb}\phantom{Ours-Single}	& 80.24	& 0.433 & 79.76 & 0.429 & 91.87	& 0.429 & 71.93 & 0.429 & 72.17 & 0.332	\\
    % \midrule
    %     Ours-CP & 81.19	& 0.429 & 81.13	& 0.425 & 91.84 & 0.425 & 73.60	 & 0.425 & 73.80	& 0.321 \\      
    %     % BERT-single + Vocab & 81.76	& 0.445 & 81.71	 & 0.443 & 92.99	 & 0.443 &73.96	 & 0.443 &74.98	& 0.303 \\
    %     \makebox[0pt][l]{Ours-Full~(CP+DE)}\phantom{Ours-Single} & \textbf{82.79} & \textbf{0.450} & \textbf{82.64}	& \textbf{0.446} & \textbf{93.46}	& \textbf{0.446} & \textbf{75.46}	& \textbf{0.446} & \textbf{76.44} 	& \textbf{0.349} \\
    \midrule
        Ours-CP & 82.03	& 0.436 & 81.86	& 0.434 & 92.46 & 0.434 & 74.71	 & 0.434 & 74.82	& 0.328 \\      
        % BERT-single + Vocab & 81.76	& 0.445 & 81.71	 & 0.443 & 92.99	 & 0.443 &73.96	 & 0.443 &74.98	& 0.303 \\
        \makebox[0pt][l]{Ours-Full~(CP+DE)}\phantom{Ours-Single} & \textbf{82.58} & \textbf{0.450} & \textbf{82.40}	& \textbf{0.447} & \textbf{92.73}	& \textbf{0.447} & \textbf{75.02}	& \textbf{0.447} & \textbf{75.73} 	& \textbf{0.354} \\
    \bottomrule
\end{tabular}
\end{adjustbox}
\caption{The experiment results on ChID. We only compute MRR for methods that have idiom embeddings.}
\label{tab:results}
\end{table*}    
% \subsection{Main Results}
\paragraph{Results:}

We show the comparison of the performance of the various methods together with the human performance in Table~\ref{tab:results}.
For Human, LM, AR and SAR, the performance shown in the table is taken directly from ChID~\cite{zheng-etal-2019-chid}.

We can observe the following from the table.
(1) In general, methods using BERT (including both the baselines and our methods) perform substantially better than previous methods based on BiLSTMs. 
This is not surprising and confirms the general observation that pre-trained BERT is generally very effective for many NLP tasks.
(2) Our two methods that use context pooling to explicitly incorporate more contextual information consistently work better than the BERT-based baselines that do not perform context pooling.
This shows the importance of using context pooling to encode long-range contextual information for the task of Chinese idiom prediction.
(3) Comparing \textbf{Ours-Full~(CP+DE)} with \textbf{Ours-CP}, we can see that \textbf{Ours-Full~(CP+DE)} consistently outperforms \textbf{Ours-CP}, 
for all evaluation splits in terms of both accuracy and MRR.
This shows that our full model using dual embeddings coupled with context-aware pooling makes the model more expressive and captures the underlying meanings of Chinese idioms better.
It is also worth noting that on the \textbf{Out} split,  \textbf{Ours-Full~(CP+DE)} achieves significant improvement over \textbf{Ours-CP}, showing better generalization ability of the dual embeddings.
% (2) Using context-aware pooling, \textbf{Ours-CP} achieves substantial gain over \textbf{BL-IdmEmb} on the more challenging split \textbf{Sim} and \textbf{Out}.
% \textbf{Ours-CP} also shows the competency in comparison with \textbf{BL-CharSeq} that without merging the passage and a candidate answer into a single sequence, we could still achieve competitive results.

It is interesting to observe that although we hypothesize that the meanings of Chinese idioms are oftentimes not compositional, \textbf{BL-CharSeq} performs better than \textbf{BL-IdmEmb~(w/o EC)}. 
We suspect that this is because the \textbf{BL-CharSeq} method allows cross attention between the passage and the characters in each candidate idiom, whereas \textbf{BL-IdmEmb~(w/o EC)} encodes both the passage and a candidate as a vectors without allowing any cross attention between them.
However, the design of \textbf{BL-IdmEmb~(w/o EC)} allows a large number of candidates to be considered, and when we use the enlarged candidate set, we see that \textbf{BL-IdmEmb} performs similarly to \textbf{BL-CharSeq}.
When we subsequently incorporate context pooling and dual embedding, we are able to achieve better performance than \textbf{BL-CharSeq}.
% It is important to note that \textbf{Ours-CP} is computationally much lighter than \textbf{BL-CharSeq} and enables us to train models by considering all idioms in the vocabulary as candidates, which is not feasible for \textbf{BL-CharSeq}.
				
Overall, we can see that the experiment results demonstrate that both context-aware pooling and dual embeddings are effective, and our proposed full method generally can outperform all the other methods we consider that represent the state of the art.

\begin{table}[t]
\centering
% \begin{adjustbox}{width=\linewidth}
% \small
\begin{tabular}{lccc}
    \toprule
        Model & Dev & Test & Out \\
    \midrule        
        Top-1~(wssb)          & 88.35 & 90.57  & \textbf{85.54} \\
        Top-2~(On The Road)   & \textbf{90.59} & \textbf{91.35} & 84.93 \\
        Top-3~(Beenle)        & 81.94 & 89.27  & 84.72 \\
     \midrule        
        BERT-base &	82.20 &	82.04 & -\\
        ERNIE-base & 82.46 &	82.28 & - \\
        % BERT-wwm-ext-meaning          & 77.08 & -  & - \\
        % BERT-wwm-ext-facial + VSL     & 79.71 & -  & - \\
        % BERT-wwm-ext-meaning + VSL    & 79.71 & -  & - \\
        % BERT-wwm-ext + VSL            & 80.87 & -  & - \\
        % BERT-wwm-ext + VSL + PCEL     & 80.88 & -  & - \\
        RoBERTa-large &	85.31 &	84.50 & - \\
        RoBERTa-wwm-large-ext &	85.81 &	85.37 & - \\
        % RoBERTa-zh-Large + VSL     & 81.63 & 81.77  & 76.34 \\
        % RoBERTa-zh-Large + VSL + PCEL     & 82.10 & 82.04  & 76.27 \\
    \midrule
    %     BERT-wwm-ext-att + OPT           & 83.03 & -  & - \\
    %     BERT-wwm-ext + OPT          & 82.89 & -  & - \\
    %     BERT-wwm-ext-facial + OPT     & 84.68 & -  & - \\
    %     BERT-wwm-ext-meaning + OPT    & 84.68 & -  & - \\
    %     BERT-wwm-ext-facial + VSL + OPT    & 87.84 & -  & - \\
    %     BERT-wwm-ext-meaning + VSL + OPT    & 87.84 & -  & - \\
    % \midrule
    %     BERT-wwm-ext + VSL + OPT          & 88.95 & -  & - \\    
    %     BERT-wwm-ext + VSL + PCEL + OPT   & 89.02 & -  & - \\   
    %     RoBERTa-zh-Large + VSL + OPT          & 89.71 & 89.12\footnotemark  &  83.77\footnotemark \\
        % RoBERTa-zh-Large + VSL + OPT   & 89.38 & 89.45  & 84.30 \\
        Ours-Full   & 89.68 & 89.55 & 84.43 \\
    \bottomrule
\end{tabular}
% \end{adjustbox}
\caption{Experiment results on ChID-Competition.}
\label{results}
    \label{tab:competition}
\end{table}

\subsection{Evaluation on ChID-Competition}

In the second set of experiments, we use \textbf{ChID-Competition}\footnote{\url{https://github.com/zhengcj1/ChID-Dataset/tree/master/Competition}}, which is the data for an online competition\footnote{\url{https://biendata.com/competition/idiom/}} on Chinese idiom comprehension.
Different from ChID, for each entry in ChID-Competition, a list of passages are provided with the same candidate set, and therefore some heuristic strategies can be used (for instance, the exclusion method).
The challenge is that ground truth answers will be similar in semantic meanings, and prediction models need to focus on their differences while comparing similar contexts to make the correct predictions.
ChID-Competition is divided into \textit{Train}, \textit{Dev}, \textit{Test} and \textit{Out} splits for different evaluation stages.

To further test the competency of our model, we evaluate the full model \textbf{Ours-Full} on \textbf{ChID-Competition}.
Considering the differences between ChID-Official and ChID-Competition, we use some heuristic methods to postprocess the predictions in order to optimize the results globally for a candidate set.
Without changing the training paradigm, we treat this problem an assignment problem during postprocessing and use Linear Sum Optimization to optimize the assignment.
% \textcolor{blue}{\textbf{Linear Sum Optimization}
The linear sum assignment problem is also known as minimum weight matching in bipartite graphs. 
The method we used is the Hungarian algorithm, also known as the Munkres or Kuhn-Munkres algorithm.
Suppose for each blank, we get a probability distribution over the candidate set $C$.
Then define a cost matrix $Z$ where $Z_{i, j}$ represents the log probability of the $i$-th blank choosing $c_{j}$.
Formally, let $X$ be a boolean matrix where $X_{i, j}$ is 1 if the $i$-th blank chooses the candidate ${j}$.
Our optimization problem can be written as
\begin{equation}
    \min \sum_i \sum_j Z_{i,j} X_{i,j},
\end{equation}
so that each candidate is assigned to at most one blank, and each blank to at most one candidate.
    
% \mhresponse{As requested by the reviewers, we add the detailed algorithm for the post-processing.}

The comparison between our method and previous methods is listed in Table~\ref{tab:competition}.
In the first section of the table, we list the top-ranked competitors from the competition leaderboard.
It is worth noting that these systems are used for competition purposes and may not be publicly available.
We then show the results using several pre-trained language models, where the results are found on the CLUE leaderboard\footnote{We show representative systems on the leaderboard as of the submission date of this paper. \url{https://github.com/CLUEbenchmark/CLUE}.}.
Finally, we list our own full model \textbf{Ours-Full}, which used a larger pre-trained RoBERTa for Chinese\footnote{\url{https://github.com/brightmart/roberta_zh}}.
The experiment results show that our full model achieves competitive results compared with the top ranked systems of the competition.

\subsection{Further Analysis Through Attribution Method}

To better understand how our models achieve consistent improvement, we adopt the gradient based attribution method, Integrated Gradients~(IG)~\cite{sundararajan2017axiomatic}, to visualize how each character contributes to the final prediction.
% We choose to attribute over the text input and our baseline of attribution is all zero vectors.
To make the visualization more readable, we first perform Chinese word segmentation to merge characters into words.
The attribution value of a word is the highest absolute value of all merged characters.

\begin{figure}[h]
\centering
\includegraphics[width=\linewidth]{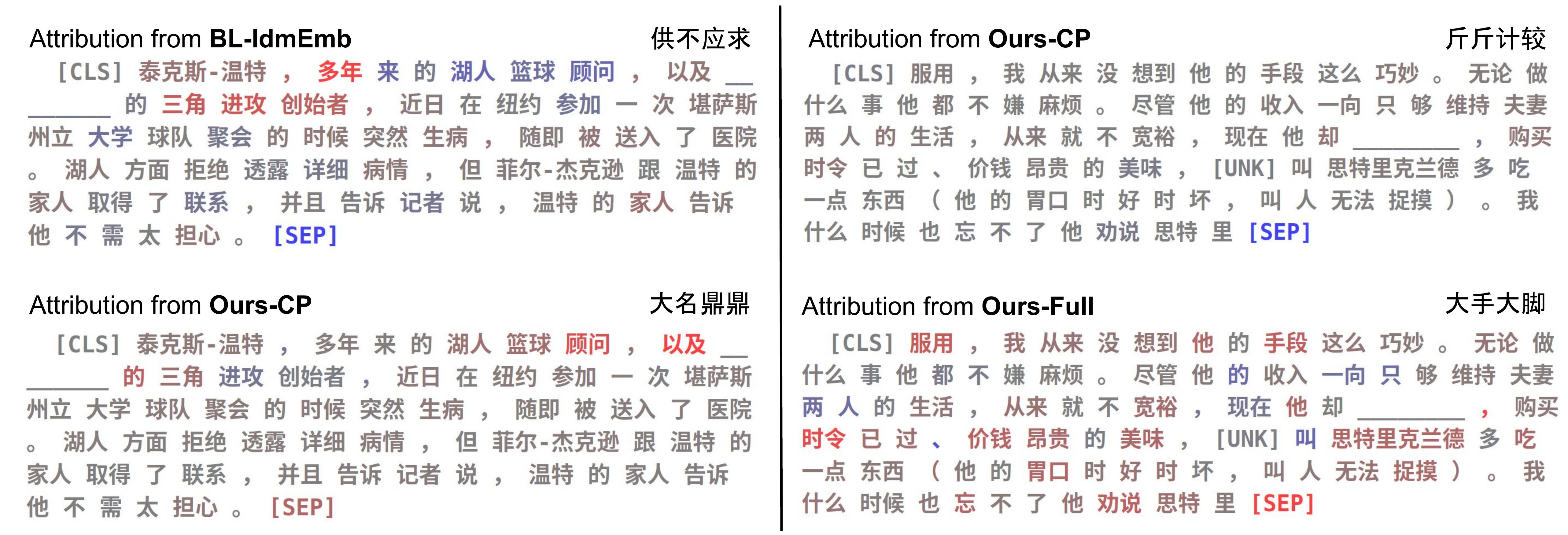}
\caption{Example cases with attribution values of words shown in red and blue. Red indicates positive correlation with the prediction while blue indicates negative correlation with the prediction.} 
\label{fig:image-text}
\end{figure}

We show some cases in Figure~\ref{fig:image-text}, where red color represents positive correlation with the prediction and blue color represents negative correlation with the prediction.
For the example on the left, both ``供不应求"~(in great demand) and ``大名鼎鼎"~(famous) are positive idioms with a sense of ``being abundant in", but the correct answer is ``大名鼎鼎" based on the context, because the context suggests that this idiom serves as an adjective to modify a person, and only ``大名鼎鼎" is used to describe a person.
On the one hand, we hypothesize that \textbf{BL-IdmEmb} may have learned the correlation between ``多年"~(for many years) and ``供不应求," and thus makes a wrong prediction solely based on this signal.
On the other hand, \textbf{Ours-CP} chooses ``大名鼎鼎", likely because it is consistent with the word ``顾问"~(consultant), which is a person, together with the conjunction word ``以及"~(and), suggesting that context-aware pooling may have helped the understanding of the context.

For the example on the right hand side of the figure, the two candidates ``斤斤计较"~(to haggle over every ounce) and ``大手大脚"~(extravagant) are antonyms and represent different attitudes towards spending money.
Both idioms suit the context well syntactically.
However, the context has the word ``却"~(but) and the word ``价钱昂贵"~(expensive), suggesting the person is extravagant with money, making ``大手大脚" the correct candidate.
This example shows that for more complex contextual understanding, \textbf{Ours-Full} has advantages over \textbf{Ours-CP}.

\section{Conclusion}
In this paper, we proposed a BERT-based dual embedding method to study Chinese idiom prediction.
We used a dual-embedding to not only capture local context information but also match the whole context passage.
% To further strengthen the learning, we consider the whole idiom vocabulary as a ranking problem.
Our experiments showed that our dual-embedding design can improve the performance of the base model, and both the idea of context-aware pooling and the idea of dual embedding can help improve the idiom prediction performance compared to the baseline methods on the ChID dataset.
% We offer insight to figurative language study that learning better representations is also important as pre-training.

% In the future, we will use attribute methods to further understand what kind of contextual words are most important for Chinese idiom prediction.

\bibliographystyle{coling}
\bibliography{coling2020}

\end{CJK*}

\end{document}